# GPT-3 Models are Few-Shot Financial Reasoners

Raul Salles de Padua, Imran Qureshi, and Mustafa U. Karakaplan

Stanford University, University of Texas at Austin, University of South Carolina

## Abstract

Financial analysis is an important tool for evaluating company performance. Practitioners work to answer financial questions to make profitable investment decisions, and use advanced quantitative analyses to do so. As a result, Financial Question Answering (QA) is a question answering task that requires deep reasoning about numbers. Furthermore, it is unknown how well pre-trained language models can reason in the financial domain. The current state-of-the-art requires a retriever to collect relevant facts about the financial question from the text and a generator to produce a valid financial program and a final answer. However, recently large language models like GPT-3 [1] have achieved state-of-the-artperformance on wide variety of tasks with just a few shot examples. We run several experiments with GPT-3 and find that a separate retrieval model and logic engine continue to be essential components to achieving SOTA performance in this task, particularly due to the precise nature of financial questions and the complex information stored in financial documents. With this understanding, our refined prompt-engineering approach on GPT-3 achieves near SOTA accuracy without any fine-tuning.

## Keywords

Question Answering, GPT-3, Financial Question Answering, Large Language Models, Infor-mation Retrieval, BERT, RoBERTa, FinQA

## 1. Introduction

Quantitative analysis is a critical tool in the financial industry. Professionals use it to ex- tract information from financial reports and make investment decisions affecting billionsof dollars every day [1]. The ability to quickly interpret the data can be the difference be- tween success and failure in today's highly competitive financial environment. Advanced experience in reasoning across structured and unstructured financial data sources and exe- cuting complicated numerical reasoning, such as comparing financial ratios of profitability or growth, are required for this type of study. These complications are exacerbated byan ever-growing quantity of financial information, making it difficult for analysts to con- duct adequate fiscal analysis and make accurate decisions. Furthermore, the vast amount of financial data gives a decisive advantage to professional investors, who can spend and hire people to manage the complexity, and creates a barrier for the average investor to participate in the market. Hence, the big question here is whether such in-depth analysis can be automated. Systems that can provide answers to financial questions would tremen- dously improve financial decision making and information transparency across all types of investors.

In this paper, we focus on Financial Question-Answering, a data analysis task that requires numerical reasoning. However, most previous research focused on the general domain, where the questions need far less calculation, such as a single-step basic math- ematical operation. In the financial sector, on the other hand, calculations can require multiple-steps and be more complicated. Therefore, Financial QA can be more difficult than traditional QA. In addition, financial QA can become problematic since the system may need to identify and grab important financial data from a variety of sources with different formats, such as tables and unstructured texts, and then develop a numerical reasoning to link all that data through calculations.

As an example, the New York Stock Exchange, which is one of many exchanges has a daily trading volume of $219bn, as of June 2022.

Page 91 from the annual reports of GRMN (Garmin Ltd.)
The fair value for these options was estimated at the date of grant using a Black-Scholes option pricing model with the following weighted-average assumptions for 2006, 2005 and 2004:

|  | 2006 | 2005 | 2004 |
|---|---|---|---|
| Weighted average fair value of options granted | $20.01 | $9.48 | $7.28 |
| Expected volatility | 0.3534 | 0.3224 | 0.3577 |
| Distribution yield | 1.00% | 0.98% | 1.30% |
| Expected life of options in years | 6.3 | 6.3 | 6.3 |
| Risk-free interest rate | 5% | 4% | 4% |

... The total fair value of shares vested during 2006, 2005, and 2004 was $9,413, $8,249, and $6,418 respectively. The aggregate intrinsic values of options outstanding and exercisable at December 30, 2006 were $204.1 million and $100.2 million, respectively. ( ... abbreviate 10 sentences ... )

Question: Considering the weighted average fair value of options, what was the change of shares vested from 2005 to 2006?
Answer: - 400
Calculations:

$$\left(\frac{9413}{20.01}\right) - \left(\frac{8249}{9.48}\right) = -400$$

Program:

divide ( 9413, 20.01 )    divide ( 8249, 9.48 )

substract ( #0, #1 )

Fig. 1. The Financial QA Task involves understanding a financial question, gathering relevant facts, and generating a program to calculate an answer. Figure from [2]

With these challenges in mind, [2] introduces FinQA, a dataset of 8,281 financial QA pairings and related numerical reasoning processes that has been annotated by experts. FinQA is created by a group of eleven financial experts based on S&P 500 earnings reports. The FinQA questions require data from tables as well as unstructured text. Many typical financial analyses computations, such as addition, comparison, and table aggregation, are used in the reasoning processes that answer these issues. They offer a retriever-generator QA architecture for retrieving supporting information from financial reports before gen- erating executable reasoning algorithms to answer the queries. Their suggested techniqueachieves an execution accuracy of roughly 60% percent when using pretrained language models such as BERT [3] and RoBERTa [4] transformer architectures. Although their method surpasses the general public (51%), the substantial accuracy difference between the model and human specialists (91%) clearly points in the direction of necessary further investigation.

In the following sections we showcase our main influential publications to this paper, present "FinQA: A Dataset of Numerical Reasoning over Financial Data" [2], our model applying a few-shot prompt-engineering using large language models, evaluation metrics, wrapping up with achieved results and analysis respectively.

## 2. PRIOR LITERATURE

Financial question answering involves two strands of research: general question answering (open QA), and financial language modelling.

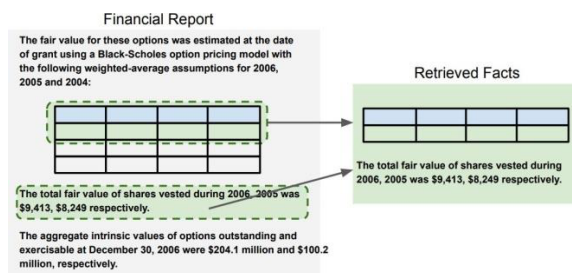

Fig. 2. FinQA Retriever Example. Figure from [2]

In open QA studies, authors attempt to combine effective performance with computa- tional efficiency. Whereas, finance related papers aim to contribute on compelling results in both domain-specific language models to semantics and downstream tasks such as sen- timent in the stock markets.

In general question answering (open QA) settings, ColBERT by [5] and ColBERT-QA by [6], tackle freezing the document encoder during training and having limited interactionwith query. Using document matrix pre-computation and late stage interaction introduced with ColBERT, ColBERT-QA finds useful passages for more questions. Improved passage relevance helps the reader component answer more accurately with greater attribution.

Baleen by [6] introduce a pipeline for multi-hop retrieval on top of ColBERT for thetask of Multi-hop QA. Multi-hop QA involves synthesizing an answer only present in twoor more documents. Baleen extends the ColBERT late interaction for this task by summa- rizing the pertinent information from retrieved passages to inform the next retrieval, and also by allowing the document matrix representations of different documents to "focus" on distinct parts of the same query. Many Multi-hop questions are multi-part complex queries. So, different documents could attend to different aspects of the query. As a re- sult of its more deliberate architecture and its stronger retrieval modeling, Baleen raises answer- recall@20 from 89% by MDR to 96% for HotPotQA benchmark finds all required passages in 92% of the examples in HoVer—up from 45% in the baseline.

RAG, by [7], investigate a general-purpose fine-tuning method for retrieval-augmented generation (RAG) — models that integrate pretrained parametric and non-parametric memory for language generation. Furthermore, RAG demonstrates state-of-the-art per- formance without separate re-ranking or reader component present in many other neural retrieval systems. They compare two RAG formulations: one that uses the same retrieved texts throughout the whole produced sequence, and the other that can utilize a different passage per token. On a variety of knowledge-intensive NLP tasks, they fine-tune and assess models, and establish the state-of-the-art on three open domain QA tasks, outper- forming parametric seq2seq models and task-specific retrieve-and-extract architectures.

Finally, there are finance domain-specific language models. One publication by [8] developed a platform for evaluating the efficacy and performance of various sentiment analysis algorithms based on a mix of text representation methods and machine learning classifiers. They run over a hundred trials using publicly available datasets that have been tagged by financial experts. They begin by evaluating particular lexicons for sentiment analysis in finance, then expand the research to encompass word and sentence encoders, all the way up to the most recent NLP transformers. Even when big datasets are not available,the results reveal that contextual embeddings outperform lexicons and fixed word and phrase encoders in sentiment analysis. Furthermore, distilled NLP transformers generate outcomes that are equivalent to their bigger teacher models, making them appropriate for usage in production situations.

Specifically for financial NLP, [9], developed a language model based on BERT called FinBERT. For two financial sentiment analysis datasets, namely TRC2-financial as a sub- set of Reuters' TRC2 and Financial PhraseBank, their results demonstrate improvementsin every measurable metric compared to existing state-of-the-art results. For financial sentiment analysis, they use two more pretrained language models, ULM-Fit and ELMo, and compare them to FinBERT. They run tests to look at a variety of elements of the model, including the impact of additional pretraining on the financial corpus, training tactics to avoid catastrophic forgetting, and fine-tuning only a small portion of model layers to reduce training time without sacrificing performance. They demonstrate that FinBERT beats state-of-the-art machine learning approaches even with a smaller training set and fine-tuning only a portion of the model.

Another work called FinBERT (BERT for Financial Text Mining), by [10], built a domain-specific language model that has been pretrained on large-scale financial corpora to address this issue. Unlike BERT, [10] build six pretraining tasks in their FinBERT that cover more knowledge and are simultaneously trained on general corpora and financial sector corpora, allowing the model to better capture linguistic knowledge and semantic information. Their FinBERT outperforms all current state-of-the-art models, according to the results.

In the general domain literature, OpenQA models differ between using generation or extractive approaches, neural retrieval methods, corpus pre-computation, query-document interaction, using a separate re-ranking process or reader. ColBERT [5] adapts deep lan- guage models from BERT and provides a highly effective and competitive setup. ColBERT-QA adopts ColBERT to OpenQA in order to handle the complexity of natural language questions and improve retrieval. Baleen's condensed retrieval architecture enhances multi-hop accuracy and robustness while learning from weak training signals. RAG proposes a model for Seq2Seq tasks with retrieval. RAG encodes documents and queries separately with a BERT encoder and uses a BART decoder to convert query and retrieved document embedding into an answer.

## 3. DATA

We use a recently released dataset, "FinQA: A Dataset of Numerical Reasoning over Financial Data" [2], which contains 8,281 financial questions. These questions ask for a numerical answer based on a unique passage associated with the question. Each question also includes the formula, called "program", which is used to generate the final answer.

FinQA is based on FinTabNet [11], a dataset comprised of publicly available earnings reports of S&P 500 companies from 1999 to 2019. The FinTabNet earnings reports contain tables, figures, and texts that outline important financial information of the companies.To adapt this dataset to the Financial QA task, FinQA applied data filtering techniquesto exclude many tables that were overly complicated. This filtered dataset of financial documents are then annotated further with a finance-related question, relevant passages, a formula used to calculate the answer, and the numerical answer itself.

Writing meaningful financial questions and annotations, however, requires specialized finance knowledge. So, the FinQA researchers recruited US-based Finance professionals to pose expert questions an for each financial document in the dataset. Separately, other finance professionals were tasked to assess the data, question quality. These assessor ex-perts reached above 90% for execution accuracy and above 85% for program accuracy with very high agreement rate (93%). Moreover, FinQA researchers also hired non-experts from Amazon Mechanical Turk to apply the same data quality procedure to verify the expert outcome. The non- experts were only able to reach about 50% for the execution and pro- gram accuracy with very low agreement rate (60%).

To summarize the FinQA dataset, there are 8,281 examples (question-answer pairs) from 2,789 pages of reports. The vocabulary consists of 22.3 thousand words. The average number of sentences in the input text is about 24. The average number of tokens in the input text is about 628. The average number of rows in the input tables is about 6. The average number of tokens in the input tables is about 59. The average number of tokensin all inputs is about 689. The maximum number of tokens in all inputs is 2,679. Finally, the average question length is about 17.

FinQA questions rely on the information found across the passage, including textand tables. The passage is broken up into sentences and tables are converted into text representations. These snippets are then used as a part of the retrieval task to identify the salient portions of a document for a given question. Around 23% of these questionsleverage information only from the text, 63% from the tables, and 14% from both text and tables.

Furthermore, each question is associated with a "program" or text-based formula that the annotator used to generate the answer. These formulas are comprised of constants, operators, numbers, and references. One example is:

$$\text{multiply(divide(60, 243), const 100)}$$

Which corresponds to the formula:

$$\frac{60}{243} \times 100$$

The most common calculations in the reasoning programs are four mathematical operations of addition, subtraction, multiplication, and division. Additions constitute about 15%, subtractions constitute about 28%, multiplications constitute about 6%, and divi- sions constitute about 45%. Together, they are about 94% of all the reasoning programs. The reason why divisions constitute the largest percentage is because financial analyses frequently rely on financial ratios. Moreover, about 59% of the calculations in FinQA are 1- step programs, about 33% are 2-step programs, and about 8% are programs with 3 stepsor more.

For our research, we have created a data filtering technique that processes valid fields where we found question-answer pairs within the dictionary keys outputting 7,228 exam-ples with train, validation, and test splits of 72%/14%/14%, respectively.

## 4. MODEL

In comparison to the retriever/generator approach, we apply a few-shot prompt-engineering approach using large language models such as GPT-3 [1].

We focus on large language models since they are the state-of-the-art, general purposemodels that can capture the large number of tokens required for few-shot financial exam- ples (which include multiple passages and tables). This versatility allows us to evaluate these models on a variety of upstream Financial QA tasks such as Fact Retrieval, and Program Generation.

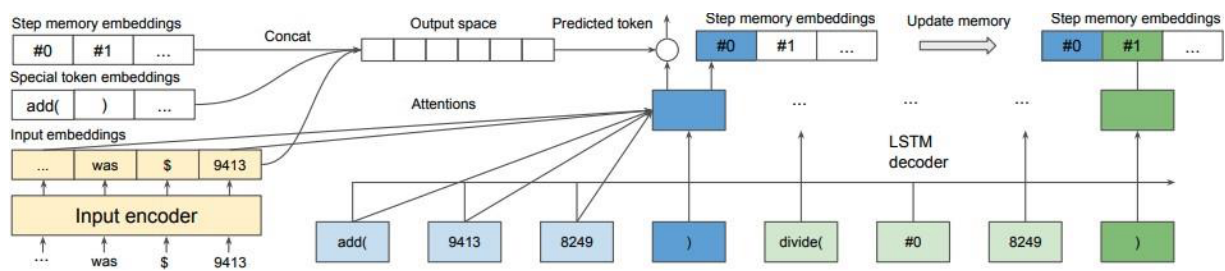

Fig. 3. Retriever-Generator Baseline Architecture for FinQA

### 4.1. Retriever/Generator Baseline with FinQA

FinQA authors [2] establish a main baseline framework called FinQANet that is built with a retriever (to get supporting facts) and a program generator (to create answers). In the former part, supporting facts are retrieved from financial reports, which surpass 2,000 tokens, to be concatenated with the question and train a classifier using pre-trained language models such as RoBERTa [4]. Top n retrieved facts will then serve as input tothe program generator. This latter part of the framework creates the program to provide answers to questions, updating the step memory token embeddings.

Our project baseline follows this approach. For the retriever, we apply a pretrained BERT- base as the classifier, where the model takes the top 3 ranked facts as retrieved results. On the generator side, we use a BERT-base [3] using adjusted FinQA open source code with the Adam optimizer [18]. Experimental details are shown in Table 3. See Figure 3for a reference architecture.

| Baselines | Exe Acc | Prog Acc |
|---|---|---|
| FinQANet (BERT) | 50 | 48 |
| FinQANet (RoBERTa) | 61.24 | 58.86 |
| Human Expert | 91.16 | 87.49 |
| General Crowd | 50.68 | 48.17 |

Table 1. FinQA baseline performance with retriever/generator models. The retriever is given in the row description, while the program generator is a trained LSTM model

## 4.2. Large Language Models

We use GPT-3 using the OpenAI API with the parameters listed in Table 2. GPT-3 isa 175B parameter autoregressive language model trained on 45TB from CommonCrawl, Wikipedia and others, showing state-of-the-art performance on many NLP-related tasks. Often these tasks can be performed by providing a relevant prompt to GPT-3, which the next predicted tokens represent the model predictions. Furthermore, language models like GPT-3 have shown numerical capabilities like addition, subtraction, division, etc. which is useful for the Financial QA task.

For our research, we trained a BERT retriever model that extracted relevant factsgiven the question and passage text. We also built a computational tree calculator which uses string parsing to execute any multi-step formula using common arithmetic operationsincluding add, subtract, multiply, divide, greater. We use both these components to evalu- ate GPT-3's financial reasoning capability. The code for both the retriever and calculator are are available publicly via Colab [12].

# 5. METHODS

## 5.1. Evaluation Metrics

Financial QA requires models to retrieve relevant facts within a document, synthesize those facts into a program, and execute that program into a numerical result. We compare retriever/generator FinQA models and pre-trained language models across various tasks using Execution Accuracy as our primary metric. Execution accuracy measures how well the model performs evaluated on the final results from the generated programs. This is effectively the average number of exact matches between the ground truth and predicted answers.

Furthermore, we look at execution accuracy within various tolerances. This metric provides insight into the directionality of our answers and whether the model is way offor is close to the ground truth answer. Effectively, accuracy tolerances allow us to proxy model understanding of the financial task at hand.

Concretely, for a given tolerance of X%, number of test examples n, ground truth values $y_i$ and predicted values $\hat{y}_i$, accuracy tolerance is calculated with the formula:

$$\frac{1}{n} \sum_{0 \leq i < n} \left(\frac{\hat{y}}{y} - 1 \leq X\right) + \left(\frac{y}{\hat{y}} - 1 \leq X\right)$$

Where A ≤ B is 1 if the inequality holds and 0 otherwise. For our analysis, we look at tolerances between 1% to 200% at 1% increments.

## 5.2. Prompt Engineering with GPT-3

We experiment with large language models by decomposing the financial QA task intoits component sub-tasks, which are: (1) Retrieve relevant passages for the question (2) Generate a computable formula with figures from the retrieved passage, and (3) Execute the computable formula into a numeric answer.

Across these tasks, we explored results by applying many-shot examples for the end- to-end tasks comprising of steps (1), (2), and (3), as well as sub-tasks by incorporatingan external retriever and a custom computational tree calculator. Each task was designed using custom functions which extract relevant information to construct a prompt which would serve as input to GPT-3.

After many trials, we found that using eight shots gave the best performance versus token tradeoff, especially given GPT3's token size limitations for retrieval and rationale tasks. For end-to-end prompting (i.e predict an answer directly from the full financial report), we used a one-shot prompt given the large token length of financial reports and the token limitations within GPT-3 models.

Furthermore, to test GPT-3's financial reasoning capability, we prompt the model with financial questions that require fixed number of steps to calculate accurately. By varying this step count, we measure GPT-3's ability to handle complex, multi-step financialquestions.

Concretely, our methods comprise of prompting GPT-3 with (a) one-shot of the full financial report, question, and corresponding ground truth numerical answer, with a test full financial report followed by a blank answer field, (b) eight-shots of the retrieved facts from financial reports with the corresponding questions and numerical answer, followed by a test retrieved facts from a financial report followed by a blank answer field, (c) eight- shots of the retrieved facts from financial reports with the corresponding questions and programs, followed by test retrieved facts from a single financial report and question with a blank program field, and (d) eight-shots of the retrieved facts from financial reports withthe corresponding program that requires N-steps, followed by test retrieved facts from a single financial report and question that requires N-steps with a blank program field. See Figures 6 and 7 for examples. For each experiment, we average results across 20 test examples.

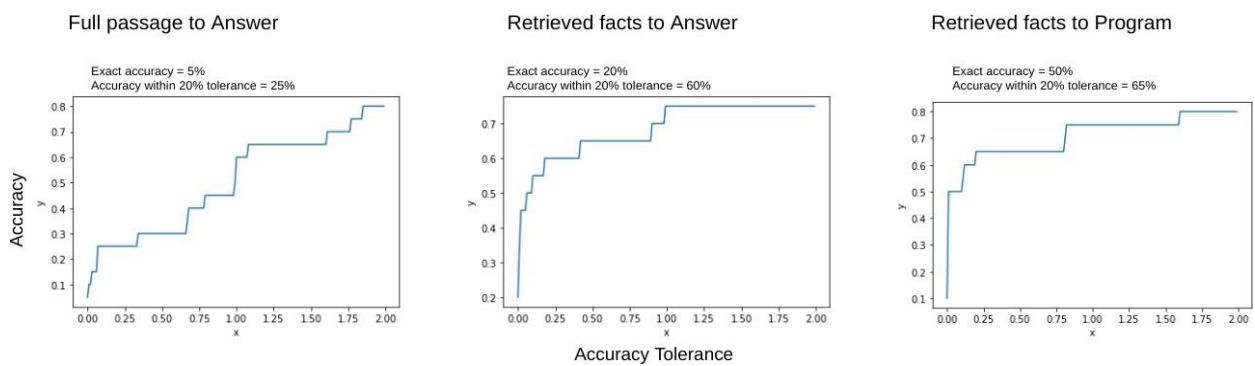

Fig. 4. Few-shot learning with GPT-3. By varying the task given to GPT3, we find that the best performance occurs when retrieval and calculation happen externally from the model. Furthermore the model becomes more directionally correct with these external components, as predictions that were not exactly right shift closer to true answers

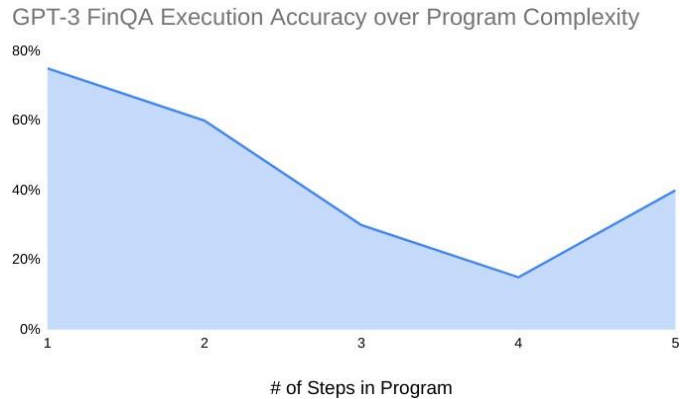

Fig. 5. GPT-3 Step Complexity: As the number of required steps to calculate the financial question increases, GPT-3 performance decreases, with a notable exception at 5 steps, where performance increases

Finally, we run our experiments with GPT-3 DaVinci, which is currently the best performing language model available publicly and so provides an upper bound on the current capabilities of pre-trained language models [1]. Details of GPT-3 hyperparameters are given in Table 2.

| Temperature | 0.1 |
|---|---|
| Max Length | 64 |
| Engine | text-davinci-002 |
| Top P | 1 |

Table 2. GPT-3 experiment details

## 6. RESULTS

When given the full passage context, the model only responded with the correct answer only 5% of the time, whereas SOTA performance is near 60%. Surprisingly, many of the GPT-3 answers were directionally correct with the correct sign, unit, and scale. See figure 4. This is reflected in the accuracy 20% tolerance which was 25%

When given the retrieved passages, GPT-3 performance increased substantially from 5% to 20%. The accuracy within at 20% tolerance also increased to 60% compared to the full- passage case.

The next experiment provided GPT-3 with retrieved passages and asked it to predict the financial program, but not execute it. In this case, the accuracy increased substantially again from 20% to 50% across many trials. There was also a slight increase in answer approximation, as accuracy within a 20% tolerance increased from 60% to 65%. See Figure8 for example results.

Finally, we compared performance across step complexity in GPT-3. As the numberof required steps to calculate the financial question increases, GPT-3 execution accuracy decreases linearly, with a notable exception at 5 steps, where accuracy increased. At 1- step, the model performed with 75% accuracy, which beats the overall SoTA accuracy and decreased linearly for 20% for the 4-step case. See 5 for more results. See Table 4 and Figure 4 in the appendix section for examples and directionality analysis.

## 7. ANALYSIS

Overall, across the various tasks mentioned in "Methods", we found that GPT-3 performs best when the retriever and calculator are handled external to the model.

While in the end-to-end experiment, GPT-3 demonstrated some capability, the poor accuracy meant that the model was struggling either to retrieve, reason about financial programs, and/or perform calculations. Interestingly, despite this poor exact accuracy, many of the answers had the correct sign and scale (e.g. if the question asked for a per- centage, the model would output a positive value < 1). This indicates that there is some latent financial reasoning capability in the model.

The retrieval experiments, where execution accuracy increased 4x, strengthened the view that GPT-3 has the capability to do financial reasoning, but struggles when asked to perform retrieval. Furthermore the results were more directionally correct, which supports the idea that GPT-3's financial reasoning was strengthened upon adding a specialized retriever model. These results also fall in line with prior retrieval enhanced model literaturesuch as as ColBERT [5], which show more efficiency by adding a retriever over the raw parameter scaling seen in GPT-3 [1].

Finally, incorporating an external calculator drove GPT-3 to near SOTA performance without any fine-tuning. The substantial increases in execution accuracy indicate that program execution is difficult for the model. This view is supported by [13], who showed that language models tend to capture statistical patterns, but not logical ones. Logical patterns, however, are required for the precise operations in financial formulae. As a result, its likely that even as language models scale, FinancialQA will continue to require a external calculator.

Generally, for program prediction, GPT-3's accuracy decreased as the complexity ofthe question increased. This is expected, and in line with human-level performance which also decreases as as the reasoning required increases. Interestingly, the 5-step accuracy increased over both 3- and 4-step cases. This is likely due to the fewer types of problems in the dataset that require a 5-step solution, over the variety of questions that exist in lower-step settings.

To summarize, we found that the large amount of information required to retrieve and generate programs made it harder for the model to generalize across retrieval, reasoning, and calculation tasks together. Furthermore, this difficulty increased linearly with the number of steps required to perform the final calculation. This indicates that the precise nature and large quantity of complex information in financial QA requires a specialized retriever and external calculator to achieve SOTA results.

## 8. CONCLUSION

Financial analysis is an important method for assessing the performance of firms. Advanced quantitative analyses are used by practitioners to answer financial queries on reports and make lucrative investment decisions. As a result, Financial Question Answering (QA) is an important question-answering task requiring in-depth numerical reasoning. A retriever must extract critical details about the financial issue from the text, and a generator must construct a legitimate computational tree and a final response. Large language models like GPT-3 have recently achieved the state-of-the-art performance on similar tasks. However, due to the nature of financial inquiries and the extensive information held in financial documents, we come up with a Financial QA system that requires a retriever and program generator. Our work produces highly promising results in terms of answers to detailed numerical financial questions that may contribute as a supporting tool to the finance industry.

Future work possibilities include running experiments with different hyperparameters and fine-tune language models to explore the limits of numerical inference within Financial QA, specially tackling the opportunity to use a transformer decoder in the generator and compare results with GPT-3.

Another path that can be explored that has the potential to add real value moving for- ward is to develop a "Conversational FinQA" dataset with human like answers to questions and context provided, that can be used to more conversational systems' developments on top of it.

**KNOWN LIMITATIONS**

Obstacles developing this publication include computational resources, as the large lan- guage models are resource-intensive, and future NLP practitioners will have to use models like GPT-3 sparingly.

Furthermore, data cleaning and filtering may pose a challenge if there are many entriesthat exceed token limits.

For this work, we focused primarily on complexity on a limited number of financial operations. Real world systems may expand those operations and should therefore be tested thoroughly to ensure similar accuracy and robustness.

**ACKNOWLEDGEMENTS**

The authors would like to thank for their families support elaborating this work as wellas Stanford University's Department of Computer Science for all the learning acquired on the areas embraced by this publication.

# REFERENCES

1. Tom Brown, Benjamin Mann, Nick Ryder, Melanie Subbiah, Jared D Kaplan, Prafulla Dhariwal, Arvind Neelakan- tan, Pranav Shyam, Girish Sastry, Amanda Askell, et al. 2020. Language models are few-shot learners. Advances in neural information processing systems, 33:1877–1901.
2. Chen, Zhiyu and Chen, Wenhu and Smiley, Charese and Shah, Sameena and Borova, Iana and Langdon, Dylan and Moussa, Reema and Beane, Matt and Huang, Ting-Hao and Routledge, Bryan and others. arXiv preprint arXiv:2109.00122. 2021. Network 13 (6) (1999) 24-30.
3. Devlin, Jacob and Chang, Ming-Wei and Lee, Kenton and Toutanova, Kristina. Bert: Pre-training of deep bidirectional transformers for language understanding. arXiv preprint arXiv:1810.04805. 2018.
4. Liu, Yinhan and Ott, Myle and Goyal, Naman and Du, Jingfei and Joshi, Mandar and Chen, Danqi and Levy, Omer and Lewis, Mike and Zettlemoyer, Luke and Stoyanov, Veselin. Roberta: A robustly optimized bert pretraining approach. arXiv preprint arXiv:1907.11692. 2019
5. Omar Khattab and Matei Zaharia. 2020. ColBERT: Efficient and Effective Passage Search via Contextualized Late Interaction over BERT, page 39–48. Association for Computing Machinery, New York, NY, USA.
6. Omar Khattab, Christopher Potts, and Matei Zaharia. 2021b. Relevance-guided Supervision for OpenQA with ColBERT. Transactions of the Association for Computational Linguistics, 9:929–944.
7. Patrick Lewis, Ethan Perez, Aleksandra Piktus, Fabio Petroni, Vladimir Karpukhin, Naman Goyal, Heinrich Küttler, Mike Lewis, Wen-tau Yih, Tim Rocktäschel, Sebastian Riedel, and Douwe Kiela. 2020. Retrieval-augmented generation for knowledge-intensive nlp tasks. In Advances in Neural Information Processing Systems, volume 33, pages 9459–9474. Curran Associates, Inc.
8. Mishev, Kostadin and Gjorgjevikj, Ana and Vodenska, Irena and Chitkushev, Lubomir T. and Tra- janov, Dimitar, journal=IEEE Access, Evaluation of Sentiment Analysis in Finance: From Lexicons to Transformers, 2020, 8, pages 131662-131682, 10.1109/ACCESS.2020.3009626
9. Dogu Araci. 2019. Finbert: Financial sentiment analysis with pre-trained language models. CoRR, abs/1908.10063.
10. Zhuang Liu, Degen Huang, Kaiyu Huang, Zhuang Li, and Jun Zhao. 2021. Finbert: A pre-trained financial language representation model for financial text mining. In Proceedings of the Twenty- Ninth International Conference on International Joint Conferences on Artificial Intelligence, pages 4513–4519.
11. Zheng, Xinyi and Burdick, Douglas and Popa, Lucian and Zhong, Xu and Wang, Nancy Xin Ru, Global table extractor (gte): A framework for joint table identification and cell structure recognition using visual context. 2021.
12. Prompt-Engineering with GPT3 - Google Colab. https://colab.research.google.com/drive/1P_QoRp-_cSZtRPSwhV0YBzkQRwqDSSQ5#scrollTo=7dQeON3i8wFb
13. Honghua Zhang, Liunian Harold Li, Tao Meng, Kai-Wei Chang, Guy Van den Broeck. On the Paradox of Learning to Reason from Data. arXiv preprint arXiv:2205.11502, 2022.
14. G.R. Blakley, Safeguarding cryptographic keys, in: Proceed- ings of the National Computer Conference, American Federation of Information, Processing Societies Proceedings, vol. 48, 1979, pp. 313-317.
15. A. Shamir, How to share a secret, Communications of the ACM 22 (1979) 612-613.
16. Seung Yi and Robin Kravetso. Moca : Mobile certificate authority for wireless ad hoc networks. In the second anunual PKI research workshop (PKI 03), Gaithersburg, 2003.
17. Dan Boneh, Ben Lynn, and Hovav Shacham (2004). "Short Signatures from the Weil Pairing". Journal of Cryptology. 17: 297-319.
18. Jimmy Ba Diederik P. Kingma. 2014. Adam: A method for stochastic optimization. arXiv preprint arXiv:1412.6980
19. Djenouri, Djamel, L. Khelladi, and N. Badache. "A survey of security issues in mobile ad hoc networks." IEEE communications surveys 7.4 (2005): 2-28.
20. Stallings, William (1990-05-03). Cryptography and Network Security: Principles and Practice. Prentice Hall. p. 165. ISBN 9780138690175.
21. H. Luo, J. Kong, P. Zerfos, S. Lu, L. Zhang, URSA: ubiquitous and robust access control for mobile ad hoc networks, IEEE/ACM Transactions on Networking 12 (6) (2004).
22. M. Narasimha, G. Tsudik, J.H. Yi, On the utility of distributed cryptography in P2P and MANETs: the case of membership control, in: Proceedings of ICNP203, 2003, pp. 336-345.

23. S. Jarecki, N. Saxena, J.H. Yi, An attack on the proactive RSA signature scheme in the URSA ad hoc network access control protocol, in: Proceedings of the SASN04, 2004, pp. 19.
24. C. Blundo, A. De Santis, A. Herzberg, S. Kutten, U. Vaccaro, M. Yung, Perfectly-secure key distribu- tion for dynamic conferences, in: Proceedings of Crypto92, LNCS, vol. 740, Springer-Verlag, 1993, pp. 471-486.
25. J. Anzai, N. Matsuzaki, T. Matsumoto, A quick group key distribution scheme with entity revocation, in: Proceedings of Asiacrypt99, LNCS, vol. 1716, Springer-Verlag, 1999, pp. 333-347.
26. V. Daza, J. Herranz, G. Sez, Constructing general dynamic group key distribution schemes with decentralized user join, in: Proceedings of ACISP03, LNCS, vol. 2727, Springer- Verlag, 2003, pp. 464-475.

**AUTHORS**

**Raul Salles de Padua** received his Master in Engineering from Instituto Tecnoĺogico de Aerońautica (ITA), from the state SP, Brazil and his Master in Business Administration from IESE Business School, from Barcelona, Spain. His Bachelor from the Universidade Federal Fluminense (UFF) in Electrical Engineering from the state of RJ, Brazil. He re- cently completed a graduate certification in Artificial Intelligence from Stanford University, where he currently serves the teaching team of Natural Language Understanding course. His research interests include Natural Language Processing and Understanding in tasks such as sentiment, question answering and text summarization.

**Imran Qreshi** is currently working at Google developing large language models for customer-specific use cases. He is currently pursuing a Masters in Computer Science from University of Texas Austin. His research interests include natural language processing and algorithms.

**Mustafa Karakaplan** holds a PhD in economics from Texas AM University. He is cur- rently a Professor of Finance at the University of South Carolina and he is specializing in artificial intelligence and machine learning at Stanford University

**Appendix**

| | |
|---|---|
| layer norm | True |
| num decoder layers | 1 |
| max_seq length | 512 |
| max_program length | 100 |
| n best size | 20 |
| dropout rate | 0.1 |
| batch size | 16 |
| batch size test | 16 |
| epoch | 100 |
| learning rate | 2e-05 |

Table 3. Retriever / Generator FinQA experiment details

| # GPT-3 | Response | Gold answer |
|---|---|---|
| 29 | 15% | 14.30% |
| 30 | 13 | 11.3 |
| 31 | 9% | 87% |
| 32 | -13% | -21% |
| 33 | 784000 | 778000 |
| 34 | 0.4 | 2.62 |
| 35 | 4.8 | 4.9 |
| 36 | 3.50% | 3.40% |
| 37 | 37.81% | 37.81 |
| 38 | 0.8 | 3.80% |

Table 4. GPT-3 DaVinci responses given one-shot passage-question-answer example compared to gold answers. Despite not being trained on the task, GPT-3 is able to extract a plausible numerical result with one example prompt.

```
passage:
entergy louisiana , inc . management's financial discussion and analysis gross operating revenues , fuel and purchased power expenses , and other regulatory credits gross operating revenues increased primarily due to : 2022 an increase of $ 98.0 million in fuel cost recovery revenues due to higher fuel rates...the industrial sector including the loss of a large industrial customer to cogeneration. .
question:
what is the growth rate in net revenue in 2003 for entergy louisiana , inc.?
answer:
5.5
passage:
investment advisory revenues earned on the other investment portfolios that we manage decreased $ 44 million , or 8.5% ( 8.5 % ) , to $ 477.8 million in 2009 . average assets in...decreased 1868 gwh in the industrial sector including the loss of a large industrial customer to cogeneration. . :
what was the average price of shares repurchased in 2010?
answer:
```

Fig.6. Example Prompt with Full Passage context

```
passage:
our cash flow metric is reconciled to the most comparable gaap measure , as follows: . ( dollars in millions ) the net cash provided by operating activities of 2013 is $ 1807...the net cash provided by operating activities of 2012 is $ 1758 ; the net cash provided by operating activities of 2011 is $ 1595 ; ( dollars in millions ) the cash flow of 2013 is $ 1170 ; the cash flow of 2012 is $ 1225 ; the cash flow of 2011 is $ 1001 ; question:
what was the average cash flow from 2011 to 2013 in millions
program:
divide(add(add(1001, add(1170, 1225)), const 3), const 2)

passage:
the weighted average interest rate under the outstanding term loans and revolving credit facility bor- rowings was 1.6% ( 1.6 % ) and 1.3% ( 1.3 % ) during the years ended december 31 , 2016 and 2015 , respectively . as of december 31 , 2016 and 2015 , the outstanding balance on the loan was $ 42.0 million and $ 44.0 million , respectively . the weighted average interest rate on the loan was 2.0% ( 2.0 question:
what is the interest expense based on the average outstanding loan balance in 2016?
program:
```

Fig.7. Example Prompt for Retrieved Passage context with program output

| | question | pred_program | gold_answer |
|---|---|---|---|
| 0 | what was the average free cash flow provided by continuing operations from 2016 to 2018 in millions | divide(add(add(1673, add(1446, 486)), const_3), const_2) | 1804.000 |
| 1 | what was the ratio of the net increase sales leading to the net increase in the operating profit in 2012 to the net decrease in the sales | divide(add(85, add(85, 75)), subtract(45, 15)) | 4.083 |
| 2 | from 2007 to 2011 what was the total expected benefits payments in thousands | add(add(add(add(117, 140), 203), 263), 328), const_5) | 1051.000 |
| 3 | how much has cash equivalents and marketable securities decreased from 2014 to 2016? | subtract(1509.7, 1100.6) | 0.339 |
| 4 | in 2009 what was the average revenues per quarter in millions | divide(add(add(243040, 245769), 248251), 258787), const_4) | 248961.750 |
| 5 | what was the average rental expense from 2007 to 2009 | divide(add(add(35614, add(39586, 42905)), const_3), const_2) | 59054.000 |
| 6 | what was the average cash outflows for real estate development investments from 2011 to 2013 | divide(add(add(162070, add(427355, 264755)), const_3), const_2) | 428.650 |
| 7 | assuming a midpoint interest rate in the range , what would be the annual interest expense on interest rate swap agreements based on the notional amounts , in millions? | divide(multiply(79.9, const_1000000), const_2) | 5.421 |
| 8 | for 2007 , what was thee average quarterly high stock price? | divide(add(add(add(41.31, 43.84), 45.45), 46.53), const_4) | 44.282 |
| 9 | what portion of the total facilities is owned by the company? | divide(add(add(add(add(add(add(41, 2), 11), 26), 1), 2), 1), 1), const_2) | 0.964 |
| 10 | what is the percentage of total debt from 2014-2015 that was long-term debt? | divide(multiply(1610.3, const_100), 1762.3) | 92.501 |

Fig.8. GPT3 Predicted Programs from Financial Questions